\documentclass[10pt,letterpaper]{article}

\usepackage{cogsci}
\usepackage{soul}
\usepackage{url}
\usepackage[hidelinks]{hyperref}
\usepackage[utf8]{inputenc}
\usepackage[small]{caption}
\usepackage{graphicx}
\usepackage{amsmath}
\usepackage{booktabs}
\usepackage{algorithm}
\usepackage{algorithmic}
\usepackage[switch]{lineno}
\usepackage[section]{placeins}
\usepackage{multirow}
\usepackage{wrapfig}
\cogscifinalcopy 

\usepackage[
  style=apa,
  natbib=true,
  annotation=false,
]{biblatex}
\usepackage{float} 

\title{Modeling Bounded Rationality in Drug Shortage Pharmacists Using Attention-Guided Dynamic Decomposition}

\author{
{\large\bfseries Yaniv Eliyahu Amiri$^1$, Noah Chicoine$^2$, Jacqueline Griffin$^2$, Stacy Marsella$^{1,3}$} \\
{\normalsize\normalfont
$^1$Khoury College of Computer Sciences, Northeastern University, Boston, MA, USA \\
$^2$Department of Mechanical and Industrial Engineering, Northeastern University, Boston, MA, USA \\
$^3$Department of Psychology, Northeastern University, Boston, MA, USA \\
\texttt{amiri.y@northeastern.edu}
}
}

\begin{document}

\maketitle

\begin{abstract}
Hospital pharmacists make high-stakes decisions to mitigate drug shortages under uncertainty, time pressure, and patient risk. Interviews revealed that pharmacists focus attention on a small subset of drugs, limiting cognitive effort to the most urgent cases. Motivated by these findings, we formalize a bounded-rational, attention-guided decision framework that dynamically decomposes drugs into a subset for high-cost reasoning and a complementary subset for low-cost monitoring. We develop two agents: an \textbf{Expert Agent} that applies attention weights derived from pharmacist interviews, and a \textbf{Learner Agent} that adapts attention allocation over time through experience. Across simulated scenarios spanning short to long horizons, we show that attention-guided planning supports stable decision-making without complete state reasoning. These results suggest that a primary decision is not what action to take, but where to allocate cognitive effort, and that attention-guided, satisficing strategies can reduce problem complexity while maintaining stable performance.

\end{abstract}

\section{Introduction}

Hospital pharmacists managing drug shortages face high-stakes sequential decisions under extreme uncertainty. Drug shortages in the United States increased steadily between 2020 and 2023, with brand-name shortages rising dramatically, creating operational and clinical risk in healthcare systems \parencite{mcgeeney2025drugshortages}. In practice, pharmacists must manage hundreds of drugs with uncertain supply, fluctuating demand, and unreliable delivery information, resulting in a partially observable high-dimensional decision problem that cannot be exhaustively planned.

Interviews with practicing pharmacists suggest that experts do not attempt global optimization across all drugs. Instead, they dynamically allocate attention to a small subset of drugs requiring immediate intervention based on their current state and risk levels, while more passively monitoring the remainder. We refer to this prioritization signal as \textbf{urgency}, which is determined by factors such as weeks of supply remaining, uncertainty in supply and demand, and prior salient events associated with each drug. Existing decision-support approaches do not explicitly model this allocation of attention. Fully rational sequential planning approaches, such as Partially Observable Markov Decision Processes (POMDPs), operate over full state spaces and become computationally intractable in realistic drug shortage settings \parencite{papadimitriou1987complexity}. Approximate planning methods improve tractability but do not capture how reasoning effort is selectively allocated under uncertainty \parencite{silver2010monte,kurniawati2008sarsop}. In interviews, experts expressed a mismatch between how these models operate and how experts reason and prioritize under time pressure. They also expressed an explainability requirement. They want these models to expose how priorities are formed or updated during decision-making so they can understand why a drug was chosen.

We therefore investigate whether selective attention can serve as a computational mechanism for tractable decision-making in these domains. We introduce two computational models. The \textbf{Expert Agent} encodes attention weights that guide which drugs to attend to. These weights are informed by pharmacist interviews. The \textbf{Learner Agent} learns these attention allocation weights over time through experience. Both agents restrict deep planning to a small urgency-based subset, reducing the number of entities considered during planning to a focused set.

We evaluate these models in simulated drug shortage scenarios informed by expert interviews and compare them to a complete state online planning baseline that considers all drugs exhaustively. Our results show that attention-guided strategies achieve stable long-term performance while substantially reducing computational cost because they determine \textit{where} to expend reasoning effort before deciding \textit{what} actions to take. These findings suggest that global optimization strategies may not be required for strong performance. We frame this model as a computational framework inspired by expert behavior.

\section{Related Work}

Hospital pharmacists must maintain a supply of hundreds of drugs in the face of a high degree of uncertainty about what their own hospital's supply is, what future demand will be, when shortages will end, and when they will be resupplied. To model the pharmacist's  decision-making process, our work builds on a range of research in formal models of human and expert decision-making and pharmaceutical supply chains.

\emph{Formal Models} Recent work in healthcare supply chains addresses drug shortages through predictive modeling \parencite{ergun2020supply}, while reinforcement learning approaches have been used to model healthcare operations \parencite{wu2023rlhealthcareops,raziei2022resilience}. These methods do not address the empirical findings that the information received by these experts is often inaccurate and subject to unpredictable changes \parencite{Chicoine2025}.

To address such uncertainty, POMDP-based models of health-center decision-making during drug shortages have been used, but their scope was limited \parencite{yongsatianchot2023agent}. A POMDP approach seeks to determine a policy that can achieve optimal solutions, a sequence of actions that maximize reward, even in the face of uncertainty about the state of the world and the effect of actions. The challenge is that finding such policies is computationally hard in the worst case \parencite{papadimitriou1987complexity}. Point-based methods \parencite{pineau2003point,kurniawati2008sarsop,silver2010monte} and deep RL-POMDP hybrids \parencite{igl2018deep} mitigate complexity by sampling or approximation, but require extensive training data and lack interpretability critical for experts to understand and trust recommendations. We sought to address these issues in our model.

\emph{Human Decision-Makers under Bounded Rationality} A different perspective comes from the study of human decision-makers, specifically the seminal work \textit{Theories of bounded rationality} \parencite{simon1972bounded}. In contrast to the standard POMDP formulation, bounded rationality argues that human problem-solving is limited due to cognitive constraints, time constraints, and imperfect information, leading to a tendency to search for solutions that "satisfice", as opposed to optimize. 

Simon's work argued additionally that expert problem solvers, in the face of bounded rationality, relied on shortcuts grounded in experience as opposed to exhaustively searching for optimal solutions \parencite{simon1972bounded}. Most notably for our work, his studies argued that expert problem-solving is a form of rapid pattern recognition that reduces the need for a step-by-step search for solutions. In short, experts see the problem differently, essentially having a different observation function, compared to novices, allowing them to simplify or eliminate the search for a solution. Recognition-Primed Decision (RPD) models decision-making in a similar way. It argues that expert decision-making is marked by the ability to recognize the salient aspects of an environment and allocate reasoning effort based on experience \parencite{klein1998sources,klein2009naturalistic}. 

\emph{Problem Decomposition and Attention} Another key approach in human problem solving is to decompose the problem into easier to solve subproblems. This addresses the cognitive constraints facing the problem solver by reducing the cognitive load. It can also facilitate the recognition of common subproblems with known solution strategies. Classical decomposition methods partition decision problems into smaller subproblems, applying the same procedure to each. Factored and structured models \parencite{poupart2005exploiting} decompose the state into weakly coupled variables but still rely on uniform procedures. 

What the problem solver is observing or attending to is closely related to the notion of decomposition. Solving a subproblem will require a shift in attention. Research on focus and attention shows that attention mechanisms in sequential decision-making improve performance in high-dimensional tasks \parencite{mnih2014recurrent,igl2018deep}. Further work \parencite{gigerenzer2011heuristic} has similarly shown that effective agents can use attention allocation strategies to direct computation to high-impact state variables. Similar work on anytime algorithms focuses on controlling when to stop computation under resource constraints, typically assuming a fixed problem representation \parencite{hansen2001monitoring}. From a cognitive perspective, this reflects a strategy of managing computation duration rather than managing what to reason about. 

Our interviews with expert pharmacists revealed a form of decomposition. They first decompose the problem by a recognition or pattern driven process that decomposes the overall problem into a set of high-risk drugs and a set of low-risk drugs. However, unlike classical decomposition, they use a non-uniform approach to solving these subproblems, devoting more cognitive and personnel resources to finding solutions in the high-risk case. This decomposition, in particular, impacts what the pharmacist focuses on, specifically reducing what the decision-maker needs to focus on first.

This decomposition into high- and low-risk drugs relates to decision theoretic work \parencite{boutilier2011decisiontheoretic} that uses non-uniformity as a layer of abstraction so that an expert can ignore certain elements of the state under certain conditions. In other words, experts focus their cognitive effort by shifting attention to certain elements of the state.

The effectiveness of this strategy hinges on recognizing which drugs require immediate attention and more extensive problem solving.

To explore this question, we use a REINFORCE-style update \parencite{williams1992simple} to adapt attention weights over urgency features. This enables the model to learn which prioritization factors support stable decision-making under uncertainty while preserving an interpretable attention mechanism.

\section{Domain Characteristics of Drug Shortage Management}

Drug shortage management is a high-dimensional, partially observable decision problem characterized by noisy signals, uncertain supply dynamics, and high-stakes outcomes. Based on semi-structured interviews with four pharmacists from different U.S. medical centers, we identify key characteristics that shape our expert-informed cognitive framework. Key signals such as true utilization, stock, and supplier communications are noisy, delayed, partially observable, and inaccurate. Supplier delivery estimates frequently change due to upstream disruptions and information asymmetries on the supplier side, with no accuracy improvement over time \parencite{Chicoine2025}. Decisions have long-horizon and path-dependent effects: actions such as implementing usage restrictions alter future demand over weeks or months. The stakes are asymmetric and high, as shortages can directly impact patient care. Pharmacists also operate under cognitive constraints, forcing them to prioritize attention toward a small subset of urgent drugs rather than exhaustively optimize over all drugs. 

\section{Example Drug Shortage Scenario}

We illustrate the resulting complexity with a representative example. Consider a hospital pharmacy managing a single class of 19 drugs.\footnote{Note: this is a simplified example, in practice, there are hundreds of oncology drugs and overall a hospital may have over a thousand drugs in active rotation} Each week, the pharmacist must decide which drugs require immediate attention and what actions to take. This decision problem involves multiple state variables, uncertain dynamics, and competing objectives.

The identified key state variables for each drug include \textbf{Quantity on Hand (QOH)}, the current inventory level, and the \textbf{Utilization Rate (UTZ)}, the weekly consumption rate. Pharmacists track \textbf{runway}, defined as the number of weeks of supply remaining (QOH divided by UTZ), which serves as the primary metric for assessing shortage risk. They simultaneously monitor supplier information including expected delivery dates and reliability.

Pharmacists take actions such as audit inventory to reduce uncertainty, implement usage restrictions called \textbf{Limited Medical Alternatives (LMAs)} \footnote{There are two LMA variants: soft LMAs reduce usage slightly, while hard LMAs are strictly enforced.}, switch between primary and alternate suppliers, contact manufacturers directly for information, or make direct purchases through reserve warehouses, hospital loan networks, or gray markets.

Each action has costs and benefits that must be weighed against an uncertain future. Audits cost staff time, but reduce uncertainty. LMAs extend runway, but may impact patient care quality. Switching suppliers may improve reliability but introduce transition risks. Emergency purchases are expensive but prevent stockouts (when supply hits zero). The challenge is that even with this example of 19 drugs and multiple possible actions per drug, the decision space becomes infeasible for exhaustive planning over a long horizon, yet pharmacists make these decisions weekly under time pressure, rarely going out of stock in necessary drugs.

Partial observability adds complexity. Pharmacists never directly observe the true state of the drug supply, but must infer it from noisy signals. The quality of the available information varies drastically as supplier signals are notoriously noisy and incomplete. Audits provide higher-confidence signals, while passive monitoring leads to increased uncertainty over time. This complexity motivates a framework that selectively allocates reasoning effort across drugs rather than planning over the full state space.

\section{Model}

The POMDP defines the underlying dynamics, while attention determines which subset of the state is considered during planning. POMDPs alone, however, are insufficient to capture how experts actually operate in this domain or the cognitive framework that they use to problem solve. Our goal is not to solve a fixed POMDP more efficiently or to approximate an optimal policy over the full state space. Instead, we address a distinct decision problem: determining which parts of the environment warrant costly reasoning at a given time.

Standard POMDP formulations implicitly assume that the full state space is always relevant for decision-making, with approximation methods reducing computational cost. Our approach introduces a selective attention mechanism. This reframes the problem from optimizing actions over a fixed state representation to controlling the allocation of computational resources across state variables. As a result, attention-guided planning goes beyond traditional optimization with the increased tractability arising from dynamic state decomposition rather than from approximation of some global objective. This reframing is not only more aligned with how experts structure complex decisions, it is also more interpretable for pharmacists in practice.

\subsection{Model Structure}

We model drug shortage management as a partially observable sequential decision problem, where the agent maintains beliefs about uncertain state variables and selects interventions under stochastic dynamics. Our framework augments this setting with an attention mechanism that determines which subset of the state is considered during planning.

\textbf{State and Belief:} The true state includes inventory (QOH), consumption rate (UTZ), runway, supplier information, action history, and consumption dynamics for each drug. Since the agent cannot observe true states directly, it maintains a belief state, tracking uncertainty over key variables.

\textbf{Action and Transition Function:} An action represents an available intervention, and the transition function describes how actions affect the state. The action space includes 12 discrete actions in five categories: monitoring, demand limits, information gathering, switching suppliers, and emergency actions. Critically, these actions differ in cost, effect, latency, uncertainty reduction, and temporal persistence. Emergency procurement actions affect inventory immediately, while usage restrictions stochastically alter future demand trajectories, and changing the active supply source can influence deliveries only after uncertain delays. Weekly dynamics are coupled across time rather than resolved in a single step, with inventory evolving according to consumption ($QOH_{t+1} = \max(0, QOH_t - UTZ_t)$) while pending actions shape future outcomes.

\textbf{Observation and Reward:} Observations are modeled with Gaussian noise to reflect real-world signal uncertainty, and uncertainty can grow over time under passive monitoring. Rewards are computed via runway-based bucketed scoring with an additional stockout penalty when inventory is depleted (or runway drops below one week), plus realistic action costs.

\subsection{Attention-Guided Planning}

For the \textbf{Expert Agent}, semi-structured interviews with hospital pharmacists were used to identify the signals they use to prioritize drugs, including runway, uncertainty in inventory and utilization, clinical impact, and prior shortage history. No direct weights or functions were specified. We then translated these reported signals into a small set of interpretable urgency components, each normalized to $[0,1]$, and used illustrative weights as a cognitive baseline. We combine these into a single attention score. For each drug $i$, \textbf{urgency} is computed as:

\begin{equation}
U_i = U_{runway}(s_i) + \sum_{f \in \mathcal{F} \setminus \{runway\}} \beta_f \cdot U_f(s_{i,f})
\end{equation}

where $U_{runway}$ provides the base urgency signal (weeks of supply remaining), and each $U_f(\cdot)$ is an urgency function mapping a drug-level belief to $[0,1]$. For each drug $i$, we compute normalized urgency components for (i) runway risk, (ii) uncertainty in runway and inventory signals, represented by the belief-state uncertainty associated with those estimates; (iii) atypical utilization level and uncertainty, calculated as the normalized deviation of a drug's utilization from the average utilization in the current scenario, together with utilization uncertainty; (iv) clinical impact (usually stable per drug); and (v) a shortage reputation term capturing prior salient shortage events. All $U_f$ terms are scaled to be comparable (values in $[0,1]$), so the attention weights $\beta_f$ control each component's contribution to the overall urgency score. Drugs with urgency $U_i$ above threshold $\tau$ are selected for focused reasoning, with a top-$k$ fallback when fewer than $k$ drugs exceed the threshold. We evaluate robustness to the choice of $\tau$ using a sensitivity analysis (Table~\ref{tab:tauablation}).

\begin{table}[t]
\centering
\small
\setlength{\tabcolsep}{4pt}
\renewcommand{\arraystretch}{0.95}
\caption{Sensitivity to urgency threshold $\tau$ (Scenario Set 3; Learner averaged over 3 seeds).}
\label{tab:tauablation}
\begin{tabular}{lrrr}
\toprule
$\tau$ & Expert reward & Learner reward & Avg.\ focus size \\
\midrule
0.55 & 58927 & 58246 & 5.1 \\
0.65 & 60025 & 58246 & 5.1 \\
0.75 & 60025 & 58704 & 5.1 \\
\bottomrule
\end{tabular}
\end{table}

\subsection{Learning Algorithm}

For the \textbf{Learner Agent}, reinforcement learning is used, but \emph{not} to learn an action-selection policy. Instead, the learner adapts the attention weights that determine which features contribute the most to urgency-based prioritization. Learning operates at the level of attention allocation. The learned component changes \textit{where} planning effort is allocated, while the same online planner is still determining \textit{what} action to take within the selected subset.

The agent maintains a vector of attention weights $\beta = (\beta_1,\ldots,\beta_{|\mathcal{F}|})$, where each $\beta_f$ scales an urgency component $U_f(\cdot)$. At each simulated week, the agent computes urgency scores for all drugs, selects a focus set using the same threshold and top-$k$ fallback rule described above, and then applies the planner only to that focused subset. After observing the outcome, the agent updates its attention weights using a REINFORCE-style surrogate gradient. Specifically, weighted feature activations are converted into softmax probabilities over drugs, and the selected focus set is used to estimate how each feature contributed to the attention decision. The resulting update increases the weights of features that were active in selected drugs when the observed reward exceeds a moving reward baseline, and decreases or stabilizes them when outcomes are worse than expected.

\begin{equation}
\label{eq:reinforce_update}
\beta_f(t+1) = \beta_f(t) + \alpha \cdot A_t \cdot \widehat{\nabla}_{\beta_f} \log P_{\text{soft}}(\text{focus}_t \mid \beta)
\end{equation}

where $A_t$ is an advantage signal from the observed reward relative to a moving reward baseline, and $\widehat{\nabla}_{\beta_f} \log P_{\text{soft}}(\text{focus}_t \mid \beta)$ is a softmax-based surrogate gradient over urgency-weighted drug scores. The threshold and top-$k$ rule define the interpretable focus set used for planning, while the softmax surrogate provides a differentiable signal for adapting attention weights.

\section{Experimental Setup and Scenario Design}

We evaluate agents in simulations of weekly drug shortage management under partial observability and stochastic supply and demand dynamics. Scenario structure and action categories follow information obtained in semi-structured interviews (weekly review cadence, runway targets, common intervention types, realistic drug scenarios), and uncertainty assumptions follow empirical findings that supplier communications and delivery estimates are noisy and unstable \parencite{Chicoine2025}. Each scenario specifies initial conditions over $n=19$ drugs, stochastic demand variation, supply disruptions, delivery delays, and noisy observations. 

We organize scenarios into three scenario sets. Scenario Set 1 (3-week horizon) tests the core intervention mechanisms under controlled disturbances. Scenario Set 2 (10-week horizon) introduces compounding effects and delayed consequences of actions like persistent demand restrictions and delayed supplier effects. Scenario Set 3 (52-week horizon) combines stressors to evaluate long-term stability of attention allocation and stockout prevention.

We compare our agents with four other agents. The \textbf{Random} agent selects actions randomly each week. The \textbf{Greedy} agent prioritizes low-runway drugs and applies immediate runway-based interventions. The \textbf{Heuristic} agent uses fixed pharmacist-inspired rules about runway, audit staleness, inventory and utilization uncertainty, supplier reliability, shipment history, and ERD uncertainty. The \textbf{Full POMDP} agent uses complete state online planning. All planning-based agents use the same underlying online planner and differ only in the subset of drugs passed to that planner, isolating whether attention-guided subset selection preserves decision quality while reducing search.
\begin{figure*}[t!]
\centering
\includegraphics[width=\textwidth]{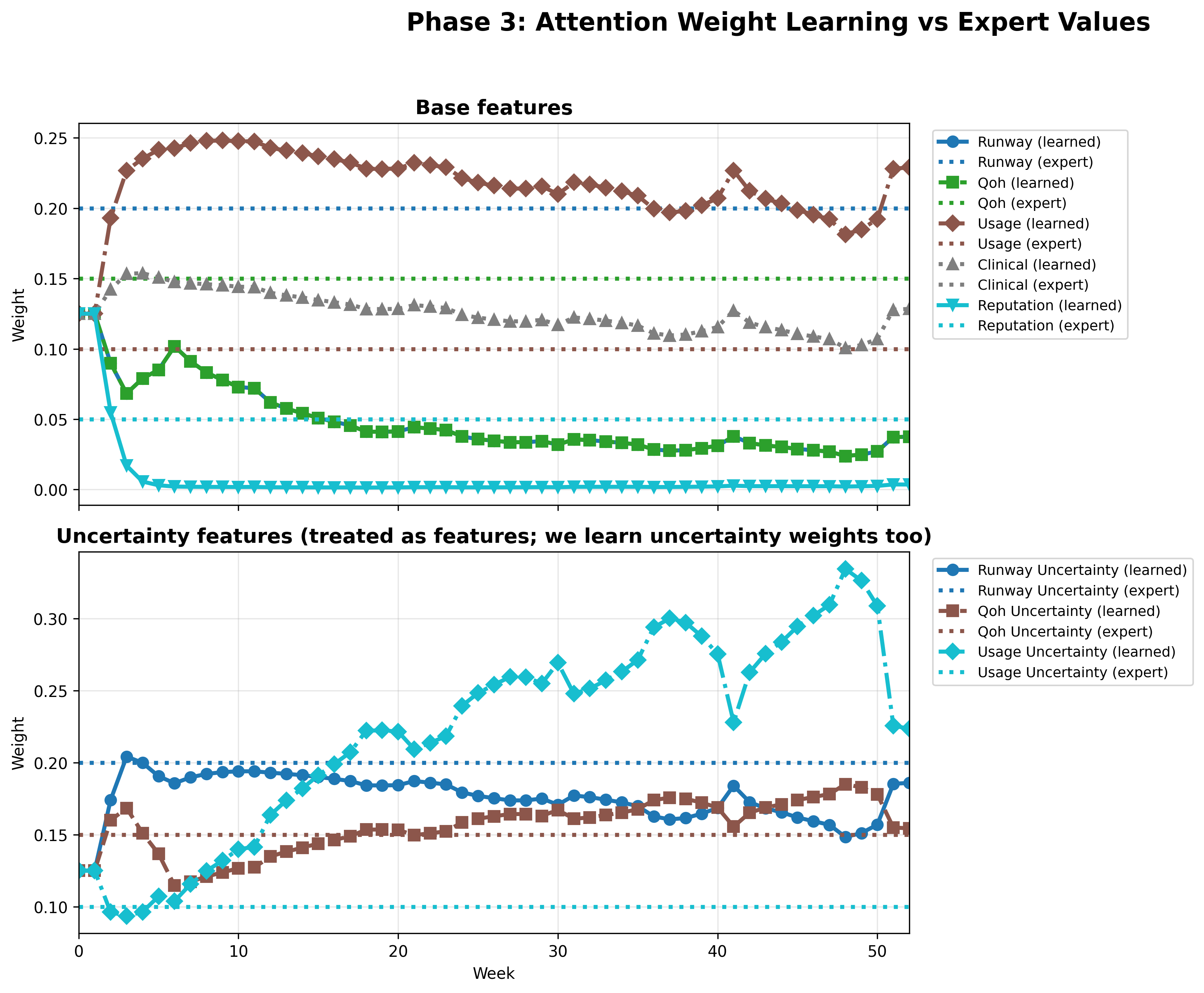}
\caption{Attention weight evolution in long-horizon scenarios. The Learner Agent adapts to the scenario by increasing weight on usage and usage uncertainty, suggesting that demand-side uncertainty becomes increasingly important for maintaining stable inventory over time.}
\label{fig:attention_learning}
\end{figure*}
\section{Results}

We report results using both decision outcomes and computational cost. Computational cost is measured as wall-clock planning time per simulated week. Our evaluation emphasizes stability and avoidance of catastrophic failures as behavioral signatures of expert reasoning rather than exact reproduction of expert behavior. Scenario Sets 1 and 2 evaluate short and medium horizons across all conditions. Scenario Set 3 evaluates long-horizon scenarios. However, the complete state planning condition is not reported for Set 3, as it becomes infeasible due to state-space explosion and minutes per simulated week.

\begin{table*}[t!]
\centering
\small
\setlength{\tabcolsep}{3.5pt}
\renewcommand{\arraystretch}{0.95}
\caption{Performance Across Short, Medium, and Long Horizon Scenarios}
\label{tab:all_phases_performance}
\begin{tabular}{llrrrrrr}
\toprule
Scenario set & Metric & Random & Greedy & Heuristic & Expert & Learner & Full POMDP \\
\midrule
\textbf{Scenario Set 1}
& Reward 
& -2,008,790 & 4,699 & 4,388 & 4,992 & \textbf{4,996} & 4,986 \\
& Time (s) 
& 0.08 & 0.08 & 0.08 & 0.28 & 0.33 & 0.59 \\
\textbf{Scenario Set 2}
& Reward 
& -1,071,850 & 874 & -1,903 & \textbf{1,684} & 1,606 & 1,605 \\
& Time (s) 
& 0.32 & 0.10 & 0.10 & 0.70 & 0.70 & 2.11 \\
\textbf{Scenario Set 3}
& Reward 
& -215,022 & 29,243 & -4,970,830 & 36,236 & \textbf{36,448} & -- \\
& Time (s) 
& 0.19 & 0.15 & 0.16 & 5.03 & 5.03 & -- \\
\bottomrule
\end{tabular}
\end{table*}

Attention-guided agents reduce planning time relative to Full POMDP in Scenario Sets 1 and 2 while maintaining comparable rewards and zero stockouts as shown in Table \ref{tab:all_phases_performance}. In Scenario Set 1, attention-guided agents require approximately 0.3 seconds per simulated week versus 0.6 seconds for Full POMDP; in Scenario Set 2, approximately 0.7 seconds versus 2.1 seconds. Gains arise because planning is restricted to high-urgency drugs.

The \textbf{Expert Agent} outperforms the naive static baselines across all scenario sets. In Scenario Set 3 long-horizon scenarios (Table \ref{tab:all_phases_performance}), the \textbf{Expert Agent} achieves 36,236 total reward compared to Greedy (29,243) and Heuristic (-4,970,830) due to the \textbf{Expert Agent}'s ability to prevent stockouts in critical realistic scenarios. The Heuristic agent experiences 4 stockouts while the \textbf{Expert Agent} and Greedy maintain zero stockouts, demonstrating the importance of POMDP-based planning with proper uncertainty reasoning. The \textbf{Learner Agent} achieves comparable outcome-level performance to the \textbf{Expert Agent} across all scenario sets while adapting its attention allocation through experience. Across Scenario Sets 1–3, the \textbf{Learner Agent} achieves similar total reward and zero stockouts (Scenario Set 1: 4,996 vs. 4,992; Scenario Set 2: 1,606 vs. 1,684; Scenario Set 3: 36,448 vs. 36,236), demonstrating that effective attention allocation can be learned online without degrading decision quality.

Figure \ref{fig:attention_learning} illustrates the learner's adaptation of attention weights over the 52-week scenario. The shift toward usage and usage uncertainty indicates that the learner increasingly prioritizes demand-side signals when maintaining stable inventory over a long horizon. complete state planning becomes infeasible in the long-horizon setting, and therefore, Scenario Set 3 should be interpreted as testing long-term stability of attention-guided and heuristic strategies rather than as a direct comparison against the complete state rational baseline (Table \ref{tab:all_phases_performance}). Both the \textbf{Expert Agent} and \textbf{Learner Agent} prevent stockouts while keeping attention bounded to a limited number of high-risk drugs. The membership of this focus set can change from week to week as urgency changes, but the number of drugs receiving detailed planning remains bounded across all long-horizon scenarios. Attention remained bounded (maximum of eight drugs) with no upward drift, indicating that learned and expert-derived attention policies support long-term stability under uncertainty.

\section{Discussion}

Our results show that attention-guided strategies can produce stable behavior in simulation. Rather than improving action selection within a fixed decision model, the proposed framework demonstrates how boundedly rational agents could determine which parts of a high-dimensional environment warrant reasoning. The \textbf{Learner Agent} further shows that attention weights can adapt while remaining stable over a long horizon.

Limitations include weekly decision cycles that may miss daily patterns, simplified supply chain dynamics, fixed POMDP-style planning assumptions, and omitted coordination dynamics such as inter-hospital collaboration, strategic supplier behavior, and decentralized decision-making. These simplifications allow us to isolate attention allocation before introducing additional organizational and temporal complexity. Future work will test the framework on more realistic scenarios, richer social dynamics, and cooperation tools for experts.

The model suggests several behavioral predictions for attention-guided decomposition. Experts should maintain a small, relatively stable focus set over time. They should attend to drugs whose runway is not yet critical when uncertainty, clinical impact, or a history of shortages make them salient. Attention should also shift with experience toward features that better predict future instability, such as utilization changes or uncertainty in supplier information. Finally, the model predicts characteristic errors: emerging risks may be missed when no factor is strong enough to push a drug into the focus set, while salient prior shortages may continue to attract attention even when runway is adequate because shortage history contributes to urgency.

\section{Conclusion}

Drug shortage pharmacists face increasing challenges in mitigating the shortages under severe uncertainty, time pressure, and risk. Through interviews with experts, we formalize attention allocation as an expert cognitive decision-structuring mechanism that determines where limited reasoning effort is applied. We introduced an attention-guided framework that partitions drugs into two subsets of differing urgency, using distinct solution strategies of online planning for a small, high-urgency focus set and lightweight monitoring for the remainder. We showcased this framework using two agents: an \textbf{Expert Agent} that encodes interview-derived attention weights and a \textbf{Learner Agent} that adapts attention weights through a REINFORCE-style update while preserving interpretability.

Across short, medium, and 52-week scenarios, the attention-guided agents achieved computational speedups relative to complete state online planning while maintaining decision quality and preventing stockouts. The \textbf{Learner Agent} further showed that effective attention allocation can adapt dynamically. These results suggest that tractability in high-dimensional, partially observable domains can be achieved by explicitly allocating computation rather than relying solely on approximating a global policy over the full state space.

Our evaluation is simulation-based and uses a fixed drug set size, and we do not claim validated scalability to hospital-wide inventories or full organizational coordination dynamics. A key next step is prospective validation with pharmacists using realistic operational data and scenario-based analyses, alongside extensions that incorporate richer demand dynamics, inter-hospital resource sharing, and incentive-aware supplier behavior.

Our findings suggest a general cognitive principle for high-stakes decision support: when the overall problem at hand is too large to plan exhaustively, the first decision is \textit{where} to spend cognitive reasoning effort.

\section{Acknowledgments}

We thank the pharmacists who participated in interviews and the NSF SAI-R drug shortages research group for their feedback and support throughout this project. This material is based upon work supported by the National Science Foundation under Grant No.~2228510, \emph{SAI-R: Designing an Improved Information Infrastructure for Better Decision Making in Pharmaceutical Supply Chains}.





\printbibliography

@techreport{mcgeeney2025drugshortages,
  author    = {McGeeney, J.D. and McAden, E. and Sertkaya, A.},
  title     = {Analysis of Drug Shortages, 2018--2023},
  institution = {U.S. Department of Health and Human Services (HHS)},
  type      = {Prepared by Eastern Research Group, Inc. for the Office of the Assistant Secretary for Planning and Evaluation (ASPE)},
  year      = {2025},
  month     = {January},
  note      = {Available from ASPE or ERG upon request}
}

@inproceedings{yongsatianchot2023agent,
  author    = {Nutchanon Yongsatianchot and Noah Chicoine and Jacqueline Griffin and Ozlem Ergun and Stacy Marsella},
  title     = {Agent-Based Modeling of Human Decision-makers Under Uncertain Information During Supply Chain Shortages},
  booktitle = {Proceedings of the 22nd International Conference on Autonomous Agents and Multiagent Systems (AAMAS 2023)},
  pages     = {1886--1894},
  year      = {2023},
  address   = {London, United Kingdom},
  publisher = {International Foundation for Autonomous Agents and Multiagent Systems (IFAAMAS)},
  url       = {https://www.ifaamas.org/Proceedings/aamas2023/pdfs/p1886.pdf}
}

@incollection{simon1972bounded,
  author    = {Herbert A. Simon},
  title     = {Theories of Bounded Rationality},
  booktitle = {Decision and Organization},
  editor    = {C.B. McGuire and Roy Radner},
  publisher = {North-Holland Publishing Company},
  address   = {Amsterdam},
  year      = {1972},
  pages     = {161--176}
}

@article{gigerenzer2011heuristic,
  title={Heuristic decision making},
  author={Gigerenzer, Gerd and Gaissmaier, Wolfgang},
  journal={Annual Review of Psychology},
  volume={62},
  pages={451--482},
  year={2011}
}

@article{papadimitriou1987complexity,
  title={The complexity of Markov decision processes},
  author={Papadimitriou, Christos H and Tsitsiklis, John N},
  journal={Mathematics of Operations Research},
  volume={12},
  number={3},
  pages={441--450},
  year={1987},
  publisher={INFORMS}
}

@inproceedings{silver2010monte,
  title={Monte-Carlo planning in large POMDPs},
  author={Silver, David and Veness, Joel},
  booktitle={Advances in Neural Information Processing Systems},
  pages={2164--2172},
  year={2010}
}

@inproceedings{kurniawati2008sarsop,
  title={SARSOP: Efficient point-based POMDP planning by approximating optimally reachable belief spaces},
  author={Kurniawati, Hanna and Hsu, David and Lee, Wee Sun},
  booktitle={Robotics: Science and Systems},
  volume={2008},
  year={2008}
}

@inproceedings{pineau2003point,
  title={Point-based value iteration: An anytime algorithm for POMDPs},
  author={Pineau, Joelle and Gordon, Geoff and Thrun, Sebastian},
  booktitle={Advances in Neural Information Processing Systems},
  volume={16},
  year={2003}
}

@article{williams1992simple,
  title={Simple statistical gradient-following algorithms for connectionist reinforcement learning},
  author={Williams, Ronald J},
  journal={Machine Learning},
  volume={8},
  number={3-4},
  pages={229--256},
  year={1992}
}

@inproceedings{mnih2014recurrent,
  title={Recurrent Models of Visual Attention},
  author={Mnih, Volodymyr and Heess, Nicolas and Graves, Alex and Kavukcuoglu, Koray},
  booktitle={Advances in Neural Information Processing Systems 27 (NIPS 2014)},
  pages={2204--2212},
  year={2014},
  publisher={Curran Associates, Inc.},
  url={https://proceedings.neurips.cc/paper_files/paper/2014/file/3e456b31302cf8210edd4029292a40ad-Paper.pdf}
}

@inproceedings{igl2018deep,
  title={Deep variational reinforcement learning for POMDPs},
  author={Igl, Maximilian and Zintgraf, Luisa and Le, Tuan Anh and Wood, Frank and Whiteson, Shimon},
  booktitle={International conference on machine learning},
  pages={2117--2126},
  year={2018},
  organization={PMLR}
}

@article{ergun2020supply,
  title={Supply chain resilience: Impact of stakeholder behavior and trustworthy information sharing with a case study on pharmaceutical supply chains},
  author={Ergun, Ozlem and Zohreh, Raziei and Atkinson, Rebekah and Keskinocak, Pinar},
  journal={International Journal of Production Economics},
  volume={227},
  pages={107663},
  year={2020},
  publisher={Elsevier}
}

@article{wu2023rlhealthcareops,
  title={Reinforcement learning for healthcare operations management: methodological framework, recent developments, and future research directions},
  author={Wu, Yifan and Wang, Zhe George and Shen, Bingqing and Zhang, Jun and Weng, Wei},
  journal={Computers \& Operations Research},
  year={2023},
  publisher={Elsevier}
}

@article{hansen2001monitoring,
  title={Monitoring and control of anytime algorithms: A dynamic programming approach},
  author={Hansen, Eric A and Zilberstein, Shlomo},
  journal={Artificial Intelligence},
  volume={126},
  number={1-2},
  pages={139--157},
  year={2001},
  publisher={Elsevier}
}

@article {Chicoine2025,
	author = {Chicoine, Noah and Griffin, Jacqueline},
	title = {The Unreliability of Estimated Release Dates in Hospital Drug Shortage Management: A Case Study of Hospital Pharmacy Operations During the COVID-19 Pandemic},
	elocation-id = {2025.07.10.25331166},
	year = {2025},
	doi = {10.1101/2025.07.10.25331166},
	publisher = {Cold Spring Harbor Laboratory Press},
	URL = {https://www.medrxiv.org/content/early/2025/07/11/2025.07.10.25331166},
	eprint = {https://www.medrxiv.org/content/early/2025/07/11/2025.07.10.25331166.full.pdf},
	journal = {medRxiv}
}

@article{boutilier2011decisiontheoretic,
  title     = {Decision-Theoretic Planning: Structural Assumptions and Computational Leverage},
  author    = {Boutilier, Craig and Dean, Thomas and Hanks, Steve},
  journal   = {arXiv preprint arXiv:1105.5460},
  year      = {2011},
  url       = {https://arxiv.org/abs/1105.5460}
}

@phdthesis{poupart2005exploiting,
  title     = {Exploiting Structure to Efficiently Solve Large Scale Partially Observable Markov Decision Processes},
  author    = {Poupart, Pascal},
  school    = {University of Toronto},
  year      = {2005},
  address   = {Toronto, Ontario, Canada},
  url       = {https://cs.uwaterloo.ca/~ppoupart/publications/ut-thesis/ut-thesis.pdf}
}

@incollection{raziei2022resilience,
  title     = {Supply Chain Resilience: Impact of Stakeholder Behavior and Trustworthy Information Sharing with a Case Study on Pharmaceutical Supply Chains},
  author    = {Raziei, Zohreh and Ergun, {\"O}zlem and Griffin, Jacqueline and Chicoine, Noah and Gong, Min and Mohaddesi, Omid and Harteveld, Casper and Kaeli, David and Marsella, Stacy},
  booktitle = {Tutorials in Operations Research: Emerging and Impactful Topics in Operations},
  pages     = {133--159},
  year      = {2022},
  publisher = {INFORMS}
}

@book{klein1998sources,
  title={Sources of Power: How People Make Decisions},
  author={Klein, Gary},
  year={1998},
  publisher={MIT Press}
}

@book{klein2009naturalistic,
  title={Streetlights and Shadows: Searching for the Keys to Adaptive Decision Making},
  author={Klein, Gary},
  year={2009},
  publisher={MIT Press}
}

\end{document}